\documentclass{article}

    \usepackage[final]{neurips_2021}

\usepackage[utf8]{inputenc} 
\usepackage[T1]{fontenc}    
\usepackage{hyperref}       
\usepackage{url}            
\usepackage{booktabs}       
\usepackage{amsfonts}       
\usepackage{nicefrac}       
\usepackage{microtype}      
\usepackage{xcolor}         

\usepackage{graphicx}
\usepackage{gensymb}
\usepackage{multirow}
\usepackage{amsmath}
\usepackage{placeins}
\usepackage{romannum}

\title{Rotation Equivariant Deforestation Segmentation and Driver Classification}

\author{%
  Joshua Mitton\thanks{https://github.com/JoshuaMitton/Rotation-Equivariant-Deforestation-Segmentation} \\
  School of Computing Science\\
  Glasgow University\\
  Glasgow, UK \\
  \texttt{j.mitton.1@research.gla.ac.uk} \\
  \And
  Roderick Murray-Smith \\
  School of Computing Science\\
  Glasgow University\\
  Glasgow, UK \\
  \texttt{roderick.murray-smith@glasgow.ac.uk} \\
}

\begin{document}

\maketitle

\begin{abstract}
Deforestation has become a significant contributing factor to climate change and, due to this, both classifying the drivers and predicting segmentation maps of deforestation has attracted significant interest. In this work, we develop a rotation equivariant convolutional neural network model to predict the drivers and generate segmentation maps of deforestation events from Landsat 8 satellite images. This outperforms previous methods in classifying the drivers and predicting the segmentation map of deforestation, offering a 9\% improvement in classification accuracy and a 7\% improvement in segmentation map accuracy. In addition, this method predicts stable segmentation maps under rotation of the input image, which ensures that predicted regions of deforestation are not dependent upon the rotational orientation of the satellite.
\end{abstract}

\section{Introduction}
Deforestation has been greatly accelerated by human activities with many drivers leading to a loss of forest area. Deforestation has a negative impact on natural ecosystems, biodiversity, and climate change and it is becoming a force of global importance \citep{foley2005global}. Deforestation for palm plantations is projected to contribute 18-22\% of Indonesia's $\mathrm{CO}_{2}$-equivalent emissions \citep{carlson2013carbon}. Furthermore, deforestation in the tropics contributes roughly 10\% of annual global greenhouse gas emissions \citep{arneth2019framing}. In addition, over one quarter of global forest loss is due to deforestation with the land being permanently changes to be used for the production of commodities, including beef, soy, palm oil, and wood fiber \citep{curtis2018classifying}. Climate tipping points are when a small change in forcing, triggers a strongly nonlinear response in the internal dynamics of part of the climate system \citep{lenton2011early}. Deforestation is one of the contributors that can cause climate tipping points \citep{lenton2011early}. Therefore, understanding the drivers for deforestation is of significant importance. 

The availability and advances in high-resolution satellite imaging have enabled applications in mapping to develop at scale \citep{roy2014landsat, verpoorter2012automated, verpoorter2014global, janowicz2020geoai, karpatne2018machine}. A range of prior works have used decision trees, random forest classifiers, and convolutional neural networks for the task of classifying and mapping deforestation drivers \citep{phiri2019long, descals2019oil, poortinga2019mapping, hethcoat2019machine, sylvain2019mapping, irvin2020forestnet}. However none of these previous methods leverage advances in group equivariant convolutional networks \citep{cohen2016group, cohen2016steerable, weiler2019general} and as such the methods are not stable with respect to transformations that would naturally occur during the capture of such data.

In this work we train models to classify drivers of deforestation and generate a segmentation map of the deforestation area. For this we build a convolutional and group equivariant convolutional model to assess the impact on classification accuracy and both segmentation accuracy and stability of the segmentation maps produced. We show that not only does the group equivariant model, with translation and rotation equivariant convolutions, improve classification and segmentation accuracy, but it has the desired property of stability of the segmentation map under natural transformations of the data capture method, namely rotations of the satellite imaging. 

\section{Equivariance}

\begin{figure}[h]
    \centering
    \begin{minipage}[t]{.34\textwidth}
        \centering
        \includegraphics[width=\linewidth]{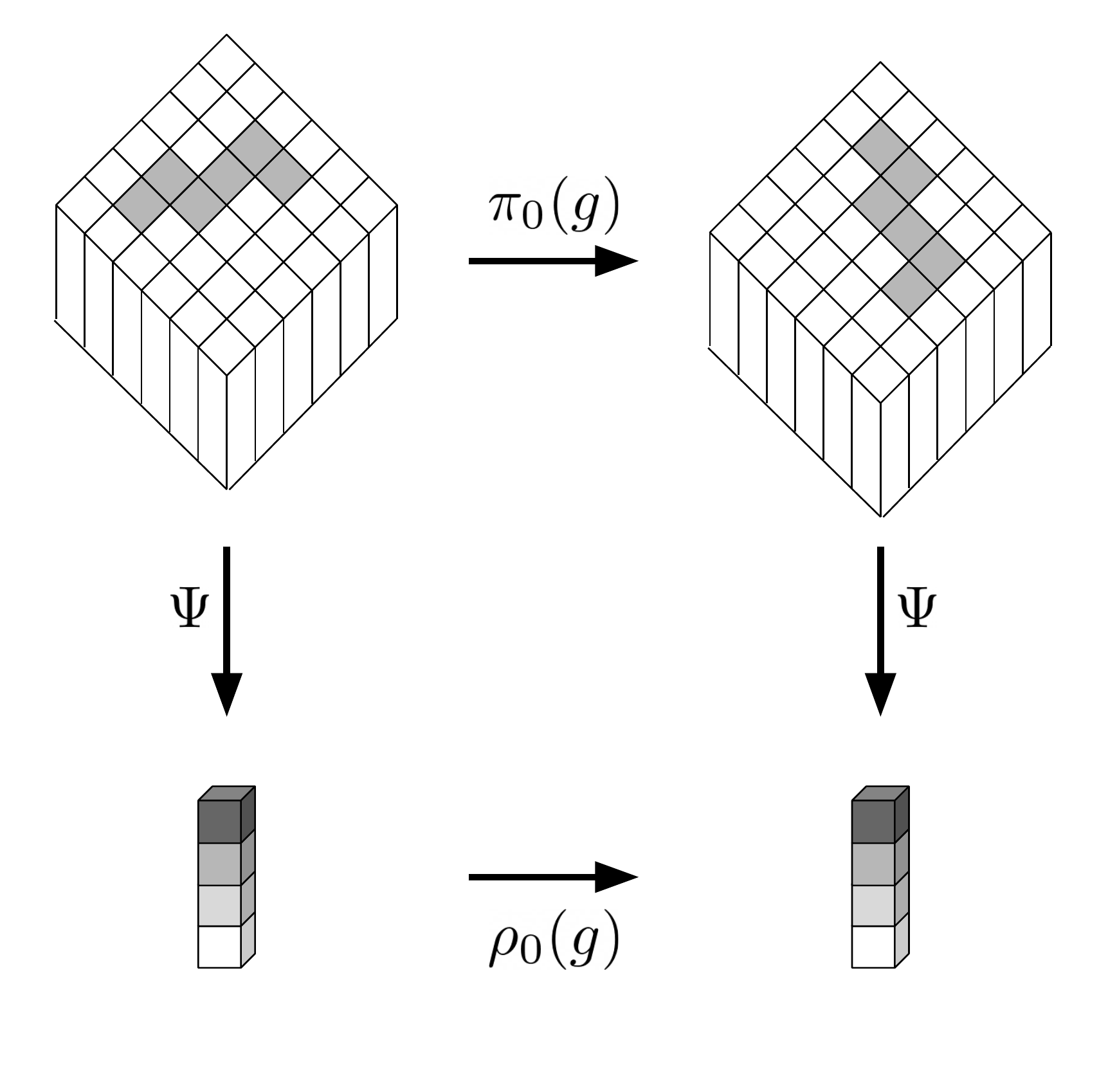}
        (a) Rotation invariant features
    \end{minipage}%
    \begin{minipage}[t]{.34\textwidth}
        \centering
        \includegraphics[width=\linewidth]{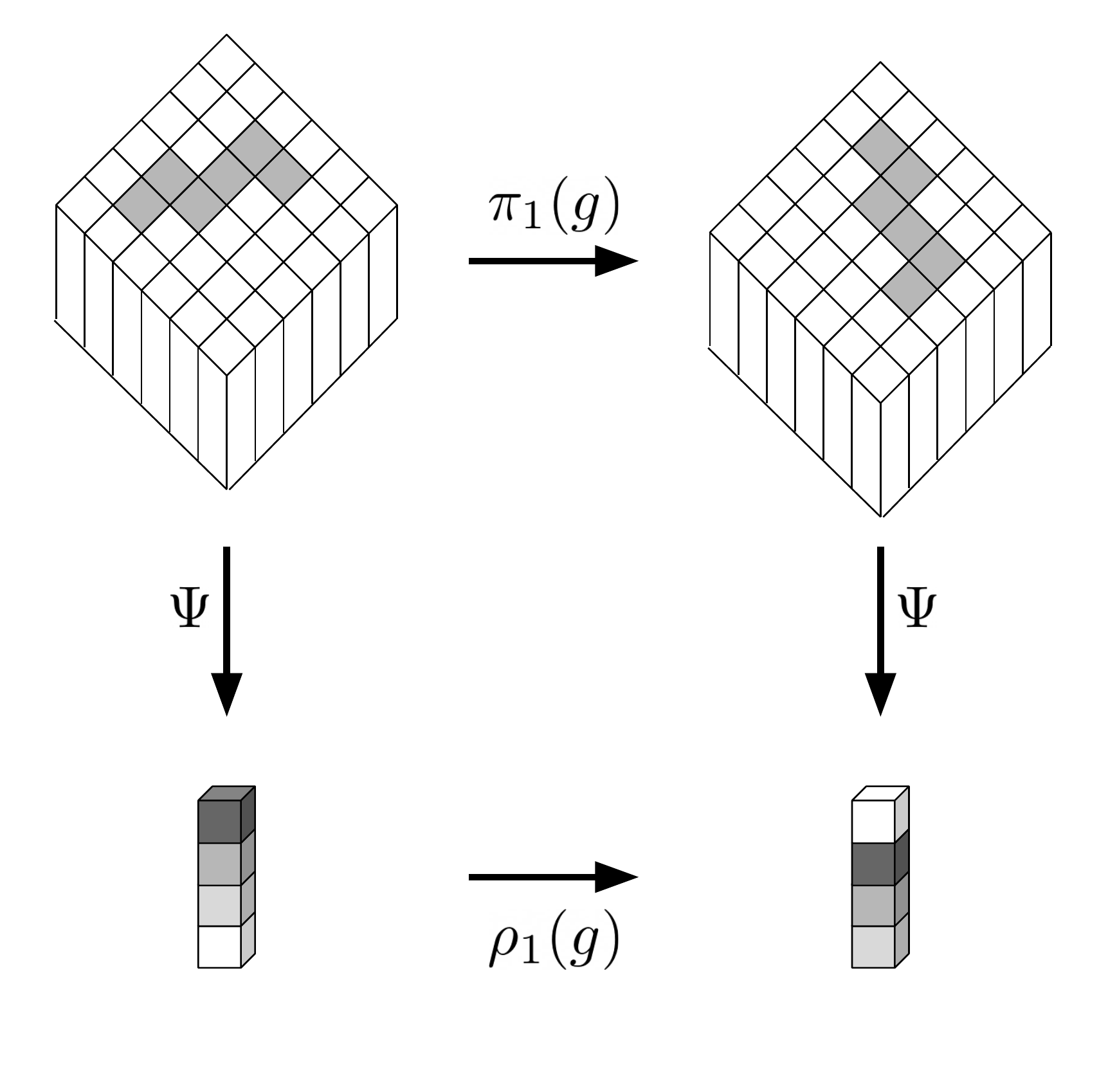}
        (b) Rotation equivariant features.
    \end{minipage}
    \caption{An image is rotated by $g$ using $\pi (g)$, where $\pi$ is some transformation law and $g$ is any angle of rotation. The filters, $\Psi$, of the layer produce some output features, here a single fiber is shown. The representation $\rho (g)$ specifies how the feature vectors transform. (a) The representation $\rho_{0}$ is the trivial representation, where $\rho_{0} (g) = 1 \; \forall g \in G$. This is used for scalar features that do not change under rotation, for example, The three RGB channels of an image are each a scalar feature and they do not mix under rotation. Therefore, typically, the input representation is the direct sum, $\bigoplus^{3}_{i=1} 1 = \mathrm{id}_{3\times3}$, of three trivial representations. (b) The representation $\rho_{1}$ is here used to represent the regular representation, where $\rho_{1} (g) = g \; \forall g \in G$. In this example the image is rotated by 90\degree which corresponds to a cyclic shift of the features in the output fiber.}
    \label{fig:equivariancereps}
\end{figure}

A CNN is, in general, comprised of multiple convolutional layers, alongside other layers. These convolutional layers are translation equivariant. This means that if the input signal is translated, the resulting output feature map is translated accordingly. Translation equivariance is a useful inductive bias to build into a model for image analysis as it is known that there is a translational symmetry within the data, i.e. if an image of an object is translated one pixel to the left, the image is still an image of the same object. This translational symmetry can be expressed through the group $\left( \mathbb{R}^{2}, + \right )$ consisting of all translations of the plane $\mathbb{R}^{2}$. 

This leads us to consider if the data has additional symmetries, such that we can look at these symmetry groups and utilise them in a model. Steerable CNNs define feature spaces of steerable feature fields $f : \mathbb{R}^{2} \rightarrow \mathbb{R}^{c}$, where a $c$-dimensional vector $f\left( x \right ) \in \mathbb{R}^{c}$ is linked to each point $x$ of the bases space \citep{cohen2016steerable}. Steerable CNNs are equipped with a transformation law that specifies how the features transform under actions of the symmetry group. The transformation law is fully characterized by the group representation $\rho$. A group representation $\rho : G \rightarrow \mathrm{GL} \left( \mathbb{R}^{c} \right )$ specifies how the channels, $c$, of the feature vector, $f\left( x \right )$, mix under transformations. For a network layer to be equivariant it must satisfy the transformation law, see Figure~\ref{fig:equivariancereps}. This places a constraint over the kernel, reducing the space of permissible kernels to those which satisfy the equivariance constraint. As the goal is to build linear layers that combine translational symmetry with a symmetry of another group for use in a model, the vector space of permissible kernels forms a subspace of that used in a conventional CNN. This increases the parameter efficiency of the layers, similar to how a CNN increases parameter efficiency over an MLP \citep{weiler2019general}. 

One particular group of interest for satellite imagery is the orthogonal group $\mathrm{O} \left( 2 \right ) = \{ O \in \mathbb{R}^{2\times2} | O^{T}O = \mathrm{id}_{2\times2} \}$. The orthogonal group consists of all continuous rotations and reflections leaving the origin invariant. In addition to the orthogonal group, the cyclic group, $\mathrm{C}_{N}$, and the dihedral group, $\mathrm{D}_{N}$, consisting of discrete rotations by angles of multiples of $\frac{2 \pi}{N}$ and in the case of the dihedral group reflections also. These rotational symmetries are of interest for analysing satellite imagery as there is no global orientation of the images collected, i.e. if an image of a forest is captured it is still the same image of the same forest if it is rotated by an angle or reflected. 

\section{Methods}

The dataset used is the same as that used by \citet{irvin2020forestnet}, where forest loss event coordinates and driver annotations were curated by \citep{austin2019causes}. Random samples of primary natural forest loss events were obtained from maps publish by Global Forest Change (GFC) at 30m resolution from 2001 to 2016. These images were annotated by an expert interpreter \citep{austin2019causes}. The drivers are grouped into categories determined feasible to identify using 15m resolution Landsat 8 imagery, while ensuring sufficient representation of each category in the dataset \citep{irvin2020forestnet}. The mapping between expert labelled deforestation driver category and driver group used as a classification target is provided in Table~\ref{driver_categories}. The dataset consists of 2,756 images, segmentation maps, and class labels; we follow the training/validation/testing set splits as provided by \citet{irvin2020forestnet}.

We use a U-Net \citep{ronneberger2015u} architecture for the task of segmentation and attach an MLP to the lowest dimensional feature space for classification. In one model we use translation equivariant convolutional layers, while in the other we use translation rotation equivariant convolutional layers. For the rotation equivariant version we choose the group $C_{8}$ of discrete rotations by $45\degree$ as the symmetry group. The input to the model is therefore three trivial representations, while hidden layers are multiple regular representations of the group, chosen similarly to the size of feature spaces in the non-rotation equivariant model, and the output is a single trivial representation. An example of how a trivial representation and a regular representation transform the output feature space is given in Figure~\ref{fig:equivariancereps}~(a) and (b) respectively. Building a model in this way will ensure that the output segmentation map is stable under $45\degree$ rotations of the input image.

\section{Results}

The model trained with rotation equivariance outperforms the non rotation equivariant model for classification of the drivers of deforestation, shown in Table~\ref{Results_classification_acc}. Given that the convolutional kernels are constrained to be rotation equivariant in the better performing model it is possible for the model to use the features more efficiently and hence model parameters are not used learning similar features at different orientations. As a result the model is able to better distinguish between the different deforestation drivers. In addition to classification accuracy, the rotation equivariant model achieves better test segmentation accuracy, demonstrated in Table~\ref{Results_segmentation_acc}. One cause of this benefit is that the model can share learned segmentation features across different orientations that occur across the different images in the dataset.


\begin{table}[h]
    \caption{Comparison between a model with translation equivariant convolutions and a model with both translation and rotation equivariant convolutions. Results are displayed as percentages for the classification accuracy of driver of deforestation.}
    \label{Results_classification_acc}
    \centering
    \begin{tabular}{lcccc}
        \toprule
        Model & Train & Validation & Test & Rotated Test \\
        \midrule
        UNET - CNN & 90.3 & 60.6 & 57.9 & 56.3 \\
        UNET - C8 Equivariant & 82.7 & 67.1 & 63.0 & 64.3 \\
        \bottomrule
    \end{tabular}
\end{table}



\begin{table}[h]
    \caption{Comparison between a model with translation equivariant convolutions and a model with both translation and rotation equivariant convolutions. Results are displayed as percentages for the segmentation accuracy of per pixel prediction averaged between the true deforestation and non-deforestation areas to account for the class imbalance towards non-deforestation areas.}
    \label{Results_segmentation_acc}
    \centering
    \begin{tabular}{lcccc}
        \toprule
        Model & Train & Validation & Test & Rotated Test \\
        \midrule
        UNET - CNN & 72.9 & 68.7 & 67.8 & 67.9 \\
        UNET - C8 Equivariant & 84.1 & 71.3 & 72.3 & 72.3 \\
        \bottomrule
    \end{tabular}
\end{table}

\begin{figure}[h]
    \centering
    \begin{minipage}[t]{\textwidth}
        \centering
        \begin{minipage}[t]{.33\textwidth}
            \centering
            \includegraphics[width=\linewidth]{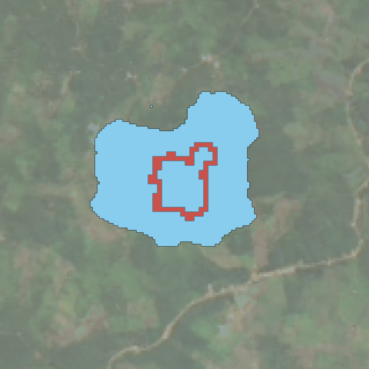}
            (\romannum{1})
        \end{minipage}
        \begin{minipage}[t]{.33\textwidth}
            \centering
            \includegraphics[width=\linewidth]{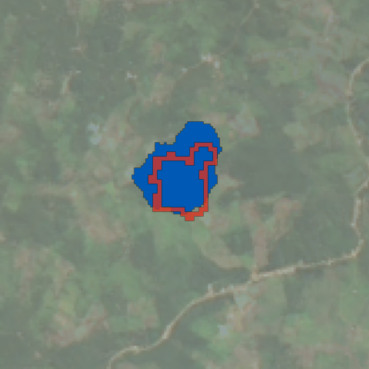}
            (\romannum{2})
        \end{minipage}
        \medskip
        \begin{minipage}[t]{.33\textwidth}
            \centering
            \includegraphics[width=\linewidth]{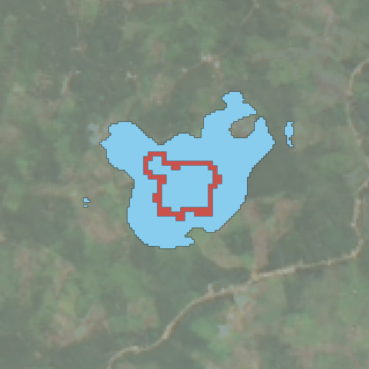}
            (\romannum{3})
        \end{minipage}
        \begin{minipage}[t]{.33\textwidth}
            \centering
            \includegraphics[width=\linewidth]{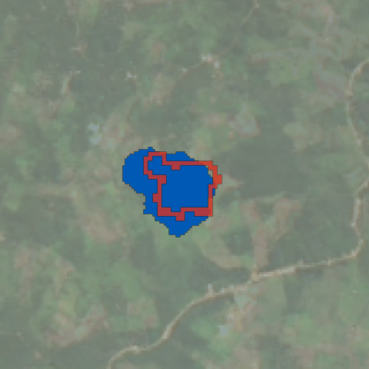}
            (\romannum{4})
        \end{minipage}
    \end{minipage}
    \caption{A comparison of predicted segmentation maps under rotation for both the non-rotation equivariant model and the rotation equivariant model. The original image is shown in (\romannum{1}) and (\romannum{2}) with the edge of the true segmentation map in red. (\romannum{1}) shows the predicted segmentation map for the non-rotation equivariant model in light blue. (\romannum{2}) shows the predicted segmentation map for the rotation equivariant model in dark blue. The $90\degree$ rotated image is shown in (\romannum{3}) and (\romannum{4}) with the edge of the true segmentation map in red. (\romannum{3}) shows the predicted segmentation map for the non-rotation equivariant model in light blue. (\romannum{4}) shows the predicted segmentation map for the rotation equivariant model in dark blue.}
    \label{fig:cnn_preds}
\end{figure}

Furthermore, the segmentation map predictions for the non rotation equivariant model and rotation equivariant models are shown to compare the stability of segmentation under rotation in Figure~\ref{fig:cnn_preds}. This highlights, in Figure~\ref{fig:cnn_preds}, that the segmentation map prediction for the non-rotation equivariant model changes as the image is rotated, which would be highly undesirable if used in practice as the rotation orientation of the satellite should not effect the segmentation map prediction of deforestation. On the other hand, the rotation equivariant models segmentation map prediction is stable under rotation, which is a desirable property of the model.

\section{Conclusion}

We develop a U-Net style model for classification and segmentation of deforestation that makes use of translation rotation equivariant convolutions. To the best of our knowledge this is the first study to make use of rotation equivariance in deforestation segmentation. The improved weight sharing through consideration of known symmetries in the data improves the classification accuracy of the model by 9\%. Furthermore, the rotation equivariant model predicts segmentation maps that are stable under rotation. In a practical application of this model this would ensure that deforestation segmentation would not be dependent upon the rotational orientation of the satellite, which does not hold true for other models. Finally, the rotation equivariant model is 7\% more accurate than the non-rotation equivariant model for the segmentation maps it produces when compared to ground truth segmentation. The improvement gain in both classification and segmentation of deforestation drivers will allow for conservation and management policies to be implemented more routinely based on model predictions from satellite data

\section{Acknowledgements}

Joshua Mitton is supported by a University of Glasgow Lord Kelvin Adam Smith Studentship. Roderick Murray-Smith acknowledges funding from the QuantIC project funded by the EPSRC Quantum Technology Programme (grant EP/MO1326X/1) and the iCAIRD project, funded by Innovate UK (project number 104690). Roderick Murray-Smith acknowledges funding support from EPSRC grant EP/R018634/1, Closed-loop Data Science.

\FloatBarrier

\bibliography{neurips_2021}
\bibliographystyle{iclr2022_conference}

\appendix

\section{Appendix}

\subsection{Dataset}

Table~\ref{driver_categories} gives the mapping between the original labels provided for the dataset by \citet{austin2019causes} and those labels used as classification targets for our models.

\begin{table}[h]
    \caption{The mapping between deforestation driver groups as defined in \citep{irvin2020forestnet} and the expert labelled deforestation driver categories defined in \citep{austin2019causes}. The deforestation driver groups are used as classification targets when training models.}
    \label{driver_categories}
    \centering
    \begin{tabular}{ll}
        \toprule
        Expert Labelled Deforestation Driver Category & Classification Target Driver Group \\
        \midrule
        Oil palm plantation & \multirow{3}{*}{Plantation} \\
        Timber plantaion &  \\
        Other large-scale plantations &  \\
        \midrule
        Grassland/shrubland & Grassland/shrubland \\
        \midrule
        Small-scale agriculture & \multirow{3}{*}{Smallholder agriculture} \\
        Small-scale mixed plantation &  \\
        Small-scale oil palm plantation &  \\
        \midrule
        Mining & \multirow{5}{*}{Other} \\
        Fish pond &  \\
        Logging road &  \\
        Secondary forest &  \\
        Other &  \\
        \bottomrule
    \end{tabular}
\end{table}

\subsection{Equivariance - Limitations and Alternative Approaches}
Equivariance places a constraint over the kernels used by the model such that the model respects symmetries in the data. An alternative approach to this is to use data augmentation, which is generally easier to implement. On the other hand, data augmentation effectively increases the size of the dataset and therefore makes training slower. Building equivariant models guarantees the models behaviour under certain symmetries, whereas data augmentation does not. Furthermore, equivariance can reduce the number of parameters required in the model and increase training efficiency. Therefore, in this work, given that we have a known symmetry group equivariant models are a sensible choice.

\subsection{Model Architecture}
For both the non-rotation equivariant and rotation equivariant models we use the same model architecture with the key difference that the convolutional layers are either rotation equivariant or non-rotation equivariant depending on the choice of model. The model architecture is a U-Net style model, which makes use of a convolutional block comprised of two convolutional layers, two batch normalisation layers, and two drop out layers. The model then consists of five convolutional blocks with downsampling in-between each and five convolutional blocks with upsampling in-between each. Further, a skip connection is placed between each convolutional block connecting upsampled layers with the corresponding same shape downsampled layer. In addition, there is a flatten layer and three multi-layer perceptron layers providing the driver classification output from the lowest dimensional space. We build the model using PyTorch \citep{NEURIPS2019_9015} and for the rotation-equivariant layers we make use of E2CNN \citep{weiler2019general}.

The non-rotation equivariant model has 3.7 million trainable parameters and the rotation equivariant model has 3.0 million trainable parameters. Each model was run on a Titan Xp GPU taking less than 30 minutes and requires approximately 3GiB of memory to train.

\end{document}